# Large Language Models for Simultaneous Named Entity Extraction and Spelling Correction

Edward Whittaker
Best Path Research Inc.
Tokyo, Japan
ed@bestpathresearch.com

Ikuo Kitagishi
Money Forward Inc.
Tokyo, Japan
kitagishi.ikuo@moneyforward.co.jp

## ABSTRACT

Language Models (LMs) such as BERT, have been shown to perform well on the task of identifying Named Entities (NE) in text. A BERT LM is typically used as a classifier to classify individual tokens in the input text, or to classify spans of tokens, as belonging to one of a set of possible NE categories.

In this paper, we hypothesise that decoder-only Large Language Models (LLMs) can also be used generatively to extract both the NE, as well as potentially recover the correct surface form of the NE, where any spelling errors that were present in the input text get automatically corrected.

We fine-tune two BERT LMs as baselines, as well as eight open-source LLMs, on the task of producing NEs from text that was obtained by applying Optical Character Recognition (OCR) to images of Japanese shop receipts; in this work, we do not attempt to find or evaluate the location of NEs in the text.

We show that the best fine-tuned LLM performs as well as, or slightly better than, the best fine-tuned BERT LM, although the differences are not significant. However, the best LLM is also shown to correct OCR errors in some cases, as initially hypothesised.

## 1 INTRODUCTION

Many business applications require the extraction of Named Entities (NE), such as names, addresses, and dates, from text data. One such application is the automatic extraction of information from printed shop receipts, which typically requires an additional Optical Character Recognition (OCR) pre-processing step. The OCR step converts a digital image of the paper receipt to machine-readable text, possibly with additional information about the coordinates of each character in the original image, from which NEs can then be extracted. This step often introduces OCR errors in the converted text representation, where visually similar characters, such as O and 0, or Y and ¥, may still be confused even when the document context is known. While most NE extraction pipelines are typically robust to handling such OCR errors, it is desirable to also recover the corrected surface form of the NE, which most existing pipelines are currently unable to do.

Recently, bi-directional, encoder-only Language Models (LMs), such as BERT[1], have been used successfully for NE extraction[1]. Typically, such LMs are used to either: (1) classify individual tokens as belonging to a particular NE category; or (2) classify a span of tokens as belonging to a particular NE category while outputting the start and end positions of the span, so that the surface form can then be recovered.

Such applications assume that the NE to be extracted appears with an identical surface form in the input text. However, when the input text contains errors, such as from text produced by performing OCR on an image, the NE category might be correct, but the NE character string itself might still contain errors. We hypothesize that using Large Language Models (LLMs)[1] generatively to output the answer to a natural language question about a receipt and a desired NE category, will generate the (potentially spell-corrected) NE, especially if it is specifically fine-tuned on such examples[2].

To investigate this hypothesis, we fine-tune two different BERT LMs, as well as eight different, open-source, Japanese-capable "base" LLMs. Each LLM has been pre-trained on different types and quantities of text data and has different numbers of model parameters. Each LLM was then fine-tuned on question-answer pairs where: (1) the question is about a particular NE category and also contains the OCR text from which the NE should be extracted; and (2) the answer contains the correct (spell-corrected where necessary) surface form of the NE. We then use each fine-tuned LLM to answer questions about each NE category of interest on the OCR'd text of each Japanese receipt in the test data, and evaluate the NE retrieval performance.

In Section 2 we describe how we fine-tune the models in this work. In Section 3 we describe the experiments that we ran and how we evaluated them. In Section 4 we report the results, and discuss and compare them in Section 5. We summarise our conclusions in Section 6 and offer suggestions for future work in Section 7.

## 2 FINE-TUNING FOR TEXT COMPLETION

Several different approaches have been proposed for supervised fine-tuning of pre-trained LLMs on a variety of downstream natural language tasks[3]. There are three main issues to decide: (1) which LLM parameters to update?; (2) what model formulation to use?;

---

[1] In this work, we collectively refer to all the models under consideration as LMs. Although BERT is sometimes considered an LLM, in this work we only refer to uni-directional, decoder-only LMs as LLMs.
[2] Note that BERT may also be fine-tuned using a masked language modelling objective, to perform spelling correction, however we do not consider such a configuration in this work.
[3] In this work, we only consider supervised fine-tuning approaches for updating an LLM's weights; we do not consider reinforcement learning based approaches, such as RLHF[4].

and (3) what data and prompt to use? We address each of these in the following sub-sections.

## 2.1 LLM adaptation using LoRA

We adopt the Low-Rank Adaptation (LoRA) method[3] which optimises only the parameters of a low-rank representation of selected sets of the original model's weights. The original model parameters are left unchanged during fine-tuning, and only the low-rank LoRA matrix parameters are optimised. These are then combined with the original weights to produce the final, fine-tuned model. LoRA has been shown to be far more memory and compute efficient than, for example, performing a full fine-tune of all parameters in the original model, while giving essentially the same fine-tuned model performance[6]. A variation on LoRA, called QLoRA, which performs the same optimisation on a quantized representation of the original LLM and has been shown to produce comparable performance to LoRA, was not used in this work so as to keep our experimental setup as straightforward as possible and minimize the number of hyper-parameters that needed to be tuned.

## 2.2 Model formulation

We formulate our task as question answering by LM text completion, employing an approach that was first described in [9][10] where an N-gram LM, trained only on the surface forms of words (i.e. no part-of-speech or parsing information), was used to answer factoid questions in the NIST TREC evaluations[4].

Given a question $Q$ we formulate the task as one of selecting the optimal answer $\hat{A}$ in the maximum likelihood sense as follows:

$$\hat{A} = \arg\max_A P(A \mid Q). \tag{1}$$

Since both $Q$ and $A$ are sequences of words (or tokens) the goal is to find the sequence of answer words $\hat{A}$ that has the highest probability given the sequence of question words, $Q$. Since $Q$ is given there is no need to score it and it only forms the context for the LM to score the answer hypotheses $A$.

Modifications to this approach were proposed in [7] which endeavoured to shape the output probability distributions so as to better match the lengths and types of answers that are expected for given question types (e.g. increasing the probability of short, date-like strings in answer to "When?" questions), based on the statistics learned from a training set of factoid question-answer pairs[5].

In [8][9][11] candidate answer word sequences were extracted from relevant texts which were obtained using keyword search and a conventional search engine[6]. Each candidate answer sequence was then scored by the LM and the top-scoring answer sequence was selected as the final answer. This approach can be interpreted as *constrained* generation where the search process is constrained to only score candidate answer sequences that were "generated" based on their frequency of occurrence in the retrieved data.

In this work, we adopt an explicitly generative approach to producing answers, where we generate the completion of the answer one token at a time, by sampling a new token from the LLM's output probability distribution until some pre-defined stopping criterion is met.

## 2.3 Prompt

What data and prompt to use for fine-tuning is closely tied to the model formulation given in Section 2.2: since we have formulated the task as sentence completion, the prompt we use should be in the form of a declarative sentence that needs completing. This also requires us to provide all the information from which to extract the answer (i.e. the OCR text from the receipt image) in the question body, along with the identity of the desired NE category for which we want the LLM to generate the answer (i.e. sentence completion).

One major issue that we encountered was that there can be significant differences in the tokenization performed by different tokenizers. For example, surrounding text can inadvertently become merged with our prompt text, which then causes issues with the Hugging Face fine-tuning code[7], when it tries to match which part of the input is the instruction, and which part of the input is the desired response.

We therefore endeavoured to compose a prompt that would work for all the tokenizers being considered, so that we could remove having different prompts as an otherwise additional variable to optimise. The final prompt used for fine-tuning of an OCR character string denoted by `{receipt}`, a desired NE category denoted by `{CAT}` and an example correct completion string denoted by `{NE}`, is shown below[8]:

" ### Question: {receipt}の{CAT}\n ### は: {NE}です。"

Note: (1) the quote characters """ are not included in the prompt; (2) the single whitespace character at the start of the prompt was found to be important since some tokenizers add special tokenizer-specific BOS and EOS tokens; and (3) the inclusion of the character sequence "です。" at the end of the prompt, which is used to indicate the termination of the LLM's answer response in a linguistically consistent manner, and also allows us to ignore any characters that may come after it in a simple post-processing step after generation[9].

The corresponding prompt for inference/generation is then simply:

" ### Question: {receipt}の{CAT}\n ### は:"

where the LLM is expected to simply complete the rest of the sentence by sampling tokens to output, up to a specified maximum number of tokens.

---

[4]A similar approach was proposed in [2], but used additional linguistic information, such as parse trees and Wordnet, and never explicitly mentioned the use of LMs.
[5]Note that there is a clear analogy here to the objective of instruction fine-tuning[12], which, instead of trying to modify the answer type distribution or the answer length distribution explicitly, modifies all parameters in the LLM to maximize the likelihood of the given answer tokens.
[6]Note that this has clear parallels to the recently popularized retrieval-augmented-generation approaches.

[7]https://huggingface.co/docs/trl/sft_trainer#train-on-completions-only
[8]The roughly equivalent English-language prompt would be a declarative sentence, like the following: " ### Question: {receipt}'s {CAT}\n ### is: {NE}."
[9]There is no obvious means to stop LLMs generating tokens; LLMs will happily continue generating tokens forever, unless we define our own stopping criterion.

While the initial " ### Question: " string is not strictly necessary, it is used by the Hugging Face code to match an instruction template[10], so it needs to be a character sequence that will be a unique and unambiguous token sequence across all the tokenizers under consideration.

On the other hand, the "\n ### は:" character sequence was found to be essential to prevent some tokenizers from merging characters from the preceding "の{CAT}", or from the following "{NE}です。" which would then cause the response template to not be found[11].

## 3 EXPERIMENTAL SETUP

### 3.1 Named entities

For the purposes of the experiments in this paper, we focus only on the six most important NE categories that we want to extract from shop receipts. Therefore, {CAT} is defined to be one of the six categories shown in Table 1. When a particular NE category does not exist in a receipt, we specify {NE} as "None" i.e. the LLM should learn to output the string "None", when no valid answer exists in {receipt}.

**Table 1: The six different NE categories considered in this work.**

| $CAT_i$ | Japanese | English |
|---|---|---|
| $CAT_1$ | 店名 | shopname |
| $CAT_2$ | 住所 | address |
| $CAT_3$ | 品目_1 | item#1 |
| $CAT_4$ | 電話番号 | telephone |
| $CAT_5$ | 日付 | date |
| $CAT_6$ | 合計 | total |

We use the character strings shown in the "Japanese" column of Table 1 to keep as much of the prompt as possible in Japanese. However, it is quite likely that using the strings in the "English" column instead would work just as well.

### 3.2 Receipt data

We have a total of 968 receipt images, each of which has been manually annotated with up to 344 different NE categories. These were split into 872 train, 18 validation and 78 test images and their corresponding annotations, as shown in Table 2.

For the train and validation data we also have manual character annotations of the surrounding text. For the test data we only have the annotated NEs, but no manually annotated information about non-NE characters. For the test and validation data we also have the text output from our OCR pipeline. We fine-tune each pre-trained LLM using three different variants of the training data set, as follows:

---

[10] Matching is *hopefully* unambiguous, but will depend on any other text that is supplied which may potentially also contain the same character sequence.
[11] Note that this sequence was found to work for our training data and the tokenizers that we investigated, but of course there is no guarantee that this will work for all possible tokenizers and input character strings. For example, if a tokenizer were to chunk together a character sequence such as "\n ### は: これ" into a single token, then the Hugging Face code would likely fail because it would no longer be possible to match the response template.

**Table 2: The number and proportion of real receipts (images and OCR text) used for training, validation and testing, and the number of characters in each partition.**

|  | Train | Validation | Test |
|---|---|---|---|
| Receipt Images | 872 | 18 | 78 |
| Percent of total | 90.1% | 1.9% | 8.0% |
| # Characters | 359,124 | 7,086 | 29,173 |

- "truth" - the original training data containing ground-truth manual character annotations of the text in each receipt image;
- "ocr1" - the "truth" dataset, but with synthetic OCR errors, where characters are corrupted randomly based on a separately computed confusion matrix; and
- "ocr10" - the same as the "ocr1" dataset, but repeated 10 times with different seeds applied each time to the random function which generates 10 different sets of training data with synthetic OCR errors.

An example receipt image from the training data is shown in Figure 1.

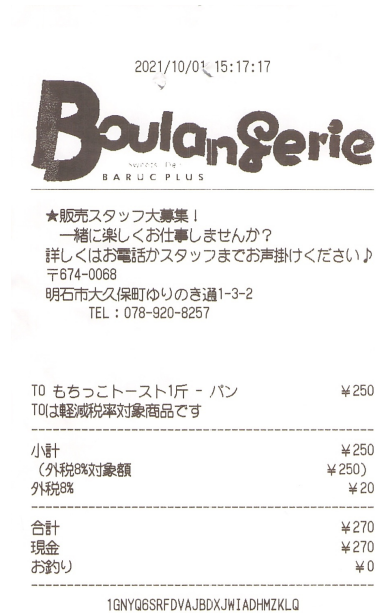

**Figure 1: An example Japanese shop receipt image.**

The corresponding sample in the "truth" training data for learning the 店名 (shopname) NE category using the manually transcribed text from the same image, is shown below:

```
 ### Question: ====2021/10/0115:17:17
Boulangerie BARUC PLUS
*販売スタッフ大募集!
-緒に楽しくお仕事しませんか?
```

```
〒674-0068明石市大久保町ゆりのき通1-3-2
TEL:078-920-8257
T0  もちっこトースト1斤-パン ¥250
T0は軽減税率対象商品です
----------------------------------------------
小計 ¥250
(外税8%対象額 ¥250)
外税8% ¥20
----------------------------------------------
合計 ¥270
現金 ¥270
お釣り ¥0
----------------------------------------------
1GNYQ6SRFDVAJBDXJWIADHMZKLQ====の店名
 ### は: Boulangerie BARUC PLUSです。
```

Similarly, the corresponding sample from the "ocr1" training data, is shown below:

```
 ### Question: ====2021/10/0115:17:17
Bou1ano*erie BARUC PLUS
*販売スタッフ大募集!
-緒に楽し<お仕車しませんか?
詳しくはお8電話かスタッフまでお戸掛けください▫
〒674-0068明石市大久保町おりのき通1-3-2
TEL:078-920-8257
T0  もちっこド-スト1斤-ン ¥250
T0は軽減税率対*象商品です
--------------------------ン---------東-------------
小計 ¥250
(外税8%対象額 ¥250)
外税8% ¥20
-----------------コ-リ---------ツ--リ--------------
合計 ¥270
現金 ¥270
お釣り ¥0
----------------------*----------------------
1%GNYQ6SRFDVAJwDXJwIADHMZKL新Q====の店名
 ### は: Boulangerie BARUC PLUSです。
```

where if you closely compare the two examples you can see that some characters have been randomly corrupted into mostly plausible mis-recognitions of the original correct characters. Also note that the answer character sequence is intentionally unaffected by these synthetic OCR errors, since our aim is for the LLM to learn to output the correct characters, irrespective of whether the input text had errors in it, or not.

### 3.3 The LLMs

We investigate the application of the two open-source BERT LMs and the eight different open-source, "base" LLMs shown in Table 3, where the name of the LM is the same as that used on the Hugging Face website. A range of "base" LMs was chosen covering a large number of different model parameters, and trained on different data. Ideally, we want to find the smallest (and therefore likely the fastest, and least compute-intensive) model that achieves the best performance.

*3.3.1 Baselines: BERT as a character classifier.* As baselines, we used (1) bert-base-multilingual-cased, the multilingual, cased BERT LM which was "pretrained on the 104 languages with the largest Wikipedias"[12], including Japanese; and (2) tohoku-nlp/bert-base-japanese-char-v3, the Japanese language-optimized, character-based BERT LM which was trained on "the Japanese portion of CC-100 dataset and the Japanese version of Wikipedia"[13]. Both models have a relatively short context of only 512 tokens, which is typically shorter than the number of characters in a single receipt from our dataset. So for both training and inference we simply split any texts that are longer than 512 characters into multiple, non-overlapping texts of maximum length 512 characters. Note that while our OCR text contains word-like units with whitespace and newlines, and each BERT LM has an associated tokenizer, we choose to process the texts at the character-level to avoid tokenization issues where an NE tag might otherwise be forced to span characters which are not actually part of the intended NE, due to the way in which the text was tokenized.

We formatted the training and validation data using the BIO format[5] (albeit without the B- tags[14]) where each character is tagged with either the neutral tag "O", a whitespace tag[15] "_", or the correspondingly annotated NE category tag of the string that the character is part of. So, in total, we used eight NE category tags for training, comprising the six NE categories used in the rest of this work, as well as the "O" and "_" tags. Moreover, we did not perform any synthetic corruption of the ground-truth training data to simulate OCR errors. The Hugging Face Tensorflow/Keras API was used for fine-tuning with a batch size of 1. Both fine-tuning and evaluation were performed entirely on a 2.4Ghz MacBook Pro 8-core Intel CPU Laptop with 64Gb RAM. We used the same data for validation as that shown in Table 2 to select the optimal point at which to terminate fine-tuning of each BERT LM, based on the character-level NE category tag classification accuracy. We then extracted NEs based on consecutive sequences of characters that were tagged with the same NE category tag. Finally, we evaluated the results in exactly the same way as for the LLMs, using the same metrics described in Section 3.4 on the 78 images of the test data for each NE category.

*3.3.2 LLMs as text generators.* Both fine-tuning and evaluation were performed on an NVIDIA A100 with 40Gb RAM using the Hugging Face PyTorch API with a batch size of 1 and 2 gradient accumulation steps. LoRA was applied to any layer in an LLM with a name that matched torch.nn.Linear, or c_attn, c_proj, and c_fc. Fine-tuning was run for 10,000 iterations for all model and data configurations and the LoRA parameters were fixed to $r = 16$, $lora\_alpha$=32, and $lora\_dropout$=0.05 for all LLMs, and a LoRA checkpoint was saved every 500 iterations. The dimensions of our investigation of the LLMs are therefore: 8 LLMs x 3 different fine-tuning datasets x 20 different LoRA checkpoints (at 20 different #iterations) x 6 different NE categories.

---

[12]https://huggingface.co/bert-base-multilingual-cased
[13]https://huggingface.co/tohoku-nlp/bert-base-japanese-char-v3
[14]Preliminary investigations using both B- and I- tags showed worse performance than only using I- tags, and also required subsequent post-processing to rectify errors such as B- tags occurring in the middle of otherwise correctly tagged NEs. In our receipt data, consecutive NEs of the same NE category are rare, so we deemed the use of B- tags to be unnecessary.
[15]In preliminary investigations, inclusion of the whitespace tag was found to be beneficial. We hypothesize that this is because varying amounts of whitespace in a shop receipt serves a similar semantic function to that of punctuation in typical prose.

For inference/generation with a fine-tuned LLM (a combination of the original LLM and one of the 20 LoRA checkpoints at $X*500$ iterations ($X = 1, 2, ...20$)), we set $temperature = 0.01$, $top\_k = 5$, $do\_sample = True$, and $max\_new\_tokens = 30$. 30 tokens was chosen to accommodate the majority of NE surface forms in our dataset, even were they to be tokenized into single characters.

For each configuration of LLM, fine-tuning data, and LoRA checkpoint, we selected the best configuration based on the validation set performance, using the same metrics described in Section 3.4, which we then evaluated on the 78 images of the test data.

### 3.4 Evaluation criteria

We calculate the precision and recall on the 18 receipts of the validation set for each of the six NE categories separately, using each configuration of fine-tuned LLM for each of the 3 different types of fine-tuning data ("truth", "ocr1", and "ocr10").

For all extracted NEs in both the ground-truth and hypotheses, we remove whitespace and punctuation before evaluation, and perform minimal character normalisation to a canonical set of visually identical characters (e.g. all full-width characters that have a half-width counterpart get normalized to the half-width character). For numerical NE categories, we remove any character in the NE that is not a number. An NE is only deemed to be a True Positive ($TP$) if the surface form exactly matches one of the similarly normalised ground-truth candidate NEs of the same NE category. If there is any mis-match after normalization, this is deemed a False Positive ($FP$). If the answer from the LLM is "None", yet a valid ground-truth answer actually exists, this is deemed a False Negative ($FN$). The LLM is always assumed to generate an answer, even if that answer is empty, or not terminated correctly with the "です。" string.

Precision and recall for each of the six NE categories ($i = 1, \ldots, 6$) are calculated in the usual manner, as follows:

$$Precision_i = \frac{\#TP_i}{\#TP_i + \#FP_i}, \quad (2)$$

$$Recall_i = \frac{\#TP_i}{\#TP_i + \#FN_i}. \quad (3)$$

In our scenario, $Precision_i$ answers the question "How many of the retrieved $i$th category of NE are correct?", whereas $Recall_i$ answers the question "How many of the $i$th category of NE that it is possible to retrieve were actually retrieved?". We assume that for each receipt, and each NE category, there is at least one valid NE (including "None") which can potentially be retrieved by the system. Moreover, our fine-tuned LLMs are always assumed to output an NE as an answer (including potentially empty strings, as well as "None") for every receipt.

For all receipts for which there is a valid ground-truth NE, we compare its surface form against the characters tagged by the BERT LMs (or generated by an LLM) based on which it is then determined to be either a $TP$, $FP$, or $FN$ (in the case that the LM output is "None", or missing, yet a ground truth answer was actually present). Note that the sum of $\#TP + \#FN$ is the total number of items that can potentially be retrieved.

The weighted F-measure for the $i$th NE category, $F_i^\beta$, is calculated as follows:

$$F_i^\beta = \frac{Precision_i \cdot Recall_i}{(\beta^2 \cdot Precision_i) + Recall_i}, \quad (4)$$

where we use $\beta = 0.5$, which assigns twice as much weight to $Precision_i$ as to $Recall_i$, and is more in line with what we prefer in our final system.

Finally, we combine the individual weighted F-measures, $F_i^\beta$, for each of the six NE categories into a single numerical measure, $F_{final}$, as follows:

$$F_{final} = \frac{\sum_{i=1}^{6} w_i F_i^\beta}{\sum_{i=1}^{6} w_i}, \quad (5)$$

where each weight $w_i$ is set equal to 1.0. We use $F_{final}^{(val)}$ to denote the weighted F-measure on the validation data, and $F_{final}^{(test)}$ to denote the weighted F-measure on the test data.

## 4 RESULTS

For each of the fine-tuned LMs, we compute $F_{final}^{(val)}$ based on the six NE categories for the 18 receipts in the validation data, and use it to select the best configuration of each fine-tuned LM on the task. This configuration is then used to compute $F_{final}^{(test)}$ using the retrieval performance of the six NE categories for the 78 receipts in the test data. These results are shown in Table 3. We also report a more detailed breakdown of the performance of the best BERT LM, the best LLM, and the smallest LLM, on each individual NE category in Table 4.

### 4.1 Baseline: BERT as a character classifier

We evaluated both fine-tuned BERT LMs using the metrics described in Section 3.4 on the 78 images of the test data. As seen in Table 3, the fine-tuned multilingual BERT LM gave the best performance of $F_{final}^{(test)} = 84.6$. A breakdown for this LM of the precision, recall and weighted F-measure values on the test set for each of the six NE categories is shown in Table 4.

### 4.2 LLMs as text generators

As seen in Table 3, the overall best LLM on the validation set was rinna/youri-7b at 5,500 iterations using the "ocr1" fine-tuning dataset. This configuration gave a final performance of $F_{final}^{(test)} = 85.6$. A breakdown of the precision, recall and weighted F-measure computed on the test set for each of the six NE categories for this model is shown in Table 4.

## 5 DISCUSSION

The OCR character error rate of our system on the test set is around 1%. However, the errors are not evenly distributed. For example, the 店名 (shopname) NE category in particular is often written using stylized fonts, which are poorly recognized by our system. Indeed, in the test set, 14 out of 78 店名 (shopname) NEs had OCR errors, 7 out of 78 品目_1 (item#1) NEs had OCR errors, and 3 out of 78 住所 (address) NEs had OCR errors. This mostly accounts for the poor extraction performance across all models of the 店名 (shopname) NE category, and the somewhat poor performance of the 品目_1 (item#1) and 住所 (address) NE categories.

Table 3: The ten LMs sorted in order of number of parameters, showing the number of trainable parameters, and $F_{final}^{(test)}$ for each fine-tuned LM configuration selected by the best $F_{final}^{(val)}$.

| LM | #Parameters | #Trainable | #Iterations | Dataset | $F_{final}^{(val)}$ ↑ | $F_{final}^{(test)}$ ↑ |
|---|---|---|---|---|---|---|
| tohoku-nlp/bert-base-japanese-char-v3 | 91M | 91M | 17x872(*)[16] | "truth" | 80.2 | 84.4 |
| cyberagent/open-calm-small | 165M | 3M | 9,500 | "truth" | 70.1 | 73.6 |
| bert-base-multilingual-cased | 177M | 177M | 12x872(†)[16] | "truth" | 83.3 | 84.6 |
| cyberagent/open-calm-medium | 409M | 7M | 4,000 | "ocr10" | 74.7 | 75.4 |
| cyberagent/open-calm-large | 840M | 10M | 4,000 | "truth" | 77.6 | 78.8 |
| line-corporation/japanese-large-lm-1.7b | 1,652M | 14M | 6,500 | "ocr10" | 78.5 | 84.5 |
| line-corporation/japanese-large-lm-3.6b | 3,556M | 24M | 10,000 | "ocr10" | 80.5 | 82.9 |
| rinna/youri-7b | 6,738M | 40M | 5,500 | "ocr1" | **85.6** | 85.6 |
| elyza/ELYZA-japanese-Llama-2-7b-fast | 6,845M | 40M | 7,500 | "ocr1" | 81.9 | 84.5 |
| stabilityai/japanese-stablelm-base-alpha-7b | 7,013M | 35M | 9,500 | "ocr1" | 82.8 | 86.0 |

Table 4: Precision, recall and weighted F-measure computed on the test set for the best performing fine-tuned BERT LM, and the best performing, as well as the smallest, LLMs on each of the six NE categories. The best $F^\beta$ for each NE category is highlighted in bold.

| | bert-base-multilingual-cased | | | rinna/youri-7b | | | cyberagent/open-calm-small | | |
|---|---|---|---|---|---|---|---|---|---|
| NE category | Precision ↑ | Recall ↑ | $F^\beta$ ↑ | Precision ↑ | Recall ↑ | $F^\beta$ ↑ | Precision ↑ | Recall ↑ | $F^\beta$ ↑ |
| 店名 (shopname) | 51.9 | 97.6 | 57.3 | 54.7 | 95.3 | **59.8** | 39.4 | 70.3 | 43.2 |
| 住所 (address) | 74.3 | 100.0 | **78.3** | 71.4 | 100.0 | 75.8 | 58.8 | 95.2 | 63.7 |
| 品目_1 (item#1) | 79.7 | 93.7 | 82.2 | 79.2 | 98.4 | **82.4** | 53.2 | 97.6 | 58.6 |
| 電話番号 (telephone) | 91.9 | 98.6 | 93.2 | 94.6 | 98.6 | **95.4** | 88.9 | 97.0 | 90.4 |
| 日付 (date) | 98.7 | 100.0 | 99.0 | 100.0 | 100.0 | **100.0** | 84.6 | 100.0 | 87.3 |
| 合計 (total) | 97.4 | 100.0 | 97.9 | 100.0 | 100.0 | **100.0** | 98.7 | 98.7 | 98.7 |

## 5.1 Baselines: BERT as a character classifier.

As expected, both BERT LMs gave good overall performance on the test data, despite having significantly fewer parameters and having been pre-trained on much less data than all but one of the LLMs. Perhaps surprisingly, bert-base-multilingual-cased gave slightly better performance than the Japanese language-optimised tohoku-nlp/bert-base-japanese-char-v3 model. We speculate that this might be because Japanese shop receipts often include a large proportion of non-Japanese text and the text itself is often somewhat idiosyncratic and quite different to the Japanese prose it had been pre-trained on.

One of the distinct advantages of the classifier approach used with the BERT LMs, over the generative approach used with the LLMs, is that it also implicitly identifies the location of the extracted NE which can then be useful for further down-stream processing where, for example, the context in which an NE was found might also be useful. On the other hand, the classification approach does not attempt to correct the surface form of the NE in the case that spelling or OCR errors are present.

## 5.2 LLMs as text generators

The results in Tables 3 and 4 show that the best fine-tuned LLM (i.e. rinna/youri-7b) outperforms the best BERT LM (i.e. bert-base-multilingual-cased) in terms of $F_{final}^{(test)}$, as well as outperforming individually on all NE categories, except 住所 (address)—although, the differences are unlikely to be statistically significant.

With the exception of the rinna/youri-7b model, we observe an interesting trend where the performance of the fine-tuned LLMs on the validation data increases with the number of model parameters. However, comparing $F_{final}^{(test)}$ of the best fine-tuned BERT LM to the LLM with the most similar number of parameters (i.e. cyberagent/open-calm-small) shows that bert-base-multilingual-cased is 14.9% relative better, while the individual performance is also better for all NE categories except the 合計 (total) NE category. While it is difficult to conclude that this is solely due to the model architecture, since there are many other differences, it is somewhat reinforced by the fact that only the largest fine-tuned LLMs, which are between 10x and 20x larger, have performance comparable to, or slightly better than, the best fine-tuned BERT LM.

---

[16]Character-level data was used to fine-tune the BERT LMs: (*) corresponds to 17 epochs of 872 receipt samples i.e. 14,824 iterations; (†) corresponds to 12 epochs of 872 receipt samples i.e. 10,464 iterations.

It is also instructive to note that the best, fine-tuned LLM was one that had been fine-tuned on the "ocr1" data, which suggests that optimising the LLM to extract an NE from an input text that might have errors is beneficial.

We would hesitate to conclude that the best, fine-tuned LLM is better than the best fine-tuned BERT LM, since they were trained on very different data and both have very different numbers of parameters and context sizes. Nonetheless, we have shown that generative LLM approaches to NE extraction can give competitive performance to state-of-the-art approaches. However, the potentially better performance of the LLMs should be considered in light of their not explicitly locating the NE in the input text, unlike the BERT LMs. This could be problematic for certain downstream tasks. A different formulation of the task that explicitly requires the LLM to learn to generate the NE together with its location in the input text might resolve this issue. This is left for future work.

In this work we have focused on NE accuracy, but the number of model parameters, and consequently the token generation speed will likely also be important factors in choosing an optimal model for a production system.

## 5.3 Correction of OCR errors

To assess whether our best fine-tuned LLM has demonstrated the ability to both generate the correct NE, as well as correct any OCR errors in the original input text, we examined the responses from the best LLM configuration on the test data. In one example of the 住所 (address) NE category, it was found that the fine-tuned LLM had correctly inserted the missing "群" in "群馬" (meaning "Gunma", a Japanese prefecture) to generate the correct address. The input OCR comprised the character sequence: "馬県高崎市栄町 1-1" which the fine-tuned LLM corrected to "**群**馬県高崎市栄町 1-1". On further investigation, it was found that this address had appeared correctly once in the fine-tuning data in another, different receipt from the same shop. It is therefore still unclear whether this ability to spell-correct came from fine-tuning or from "knowledge" contained in the pre-trained LLM. A second, more minor correction was also observed, where the fine-tuned LLM added the missing "F" (indicating floor) to the end of the address, thus correcting the address in the input OCR text from: "東京都立川市柴崎町 3-2-1 グランデュオ立川 7" to "東京都立川市柴崎町 3-2-1 グランデュオ立川 7**F**".

Several similar examples were found for 店名 (shopname), in which OCR errors had caused the insertion of incorrect characters at the beginning of several shop names. These were successfully ignored by the best fine-tuned LLM, but we consider that this is more a property of the effective NE generation than an ability to spell-correct per se.

Strangely, the best fine-tuned LLM sometimes hallucinated the wrong answer, even though the OCR input text contained the correct character string. For example, "入浴" was extracted instead of "入泉" from "No0\n 入泉\n2 個 X 単 500 ¥1,000\n", "KDonald" instead of "MisterDonut" from "O mister\nK Donut\n", and "モンシェ-ル" instead of "モンロワ-ル" from "モンロワ-ル麻布上番店".

## 5.4 Correction of manual text corruptions

To get a better feel for what our best fine-tuned LLM was capable of, we selected the OCR text for one receipt from the test set and manually corrupted it to determine what kinds of character corruptions the LLM is able to correct. The following is a receipt from "FamilyMart" (receipts from the same convenience store chain appear 39 times in our fine-tuning data):

```
<s>   ### Question: E FamilyMart
麻布十番商店街通り店
東京都港区麻布十番2丁目2-4
電話:03-5439-6226
**水曜・土曜はポイント2倍**
キャンペ-ン詳細・エントリ-は
【ファミマ 水土2倍】で検索!!
2021年 3月15日(月)20:56
レジ 2-2145 責No. 735
領 収 証
ス-パ-ドライ500缶
承 @286x 2点 ¥572
ウィルキンソン 1L ¥149軽
オ-ルフリ-350 ¥147軽
ドライゼロ350 ¥147軽
合 計 ¥1,015
(10%対象 ¥572)
( 8%対象 ¥443)
(内消費税等 ¥84)
QUICPay支払 ¥1,015
「軽」は軽減税率対象商品です。
・・・・・・・・・・・・・・・・・・・・・・・・・・・・・・
QUICPay支払
利用日時 2021/03/15 20:56:20
伝票番号 -  02145
カ-ドID X4223
有効期限 XX/XX
承認番号 0002256
取引内容 売上
支払区分 -括
支払金額 ¥1,015
・・・・・・・・・・・・・・・・・・・・・・・・・・・・・・
お客様控え
の{CAT}
 ### は:
```

where {CAT} is replaced with "住所" or "店名" as appropriate in the following explanation.

We first began by examining how far we could corrupt the characters in "FamilyMart" yet still retrieve the correct surface form for the 店名 (shopname) NE category. After trying various individual character corruptions and deletions, we realised that even if we deleted the entire "FamilyMart" character sequence, the best fine-tuned LLM was still able to return the correct answer. However, when we also deleted "ファミマ" (a Japanese abbreviation for "FamilyMart"), which appears in the 7th line of the receipt, the LLM returned the 品目_1 (item#1) NE category instead. We speculate that the LLM had learned "ファミマ" as an additional feature, and associated it with the correct 店名 (shopname) NE category, so that when it was also missing, it opted for the next most likely character string instead.

We then turned our attention to returning the 住所 (address) NE category from the same receipt. In the original OCR text, the character sequence containing the address did not contain any OCR errors, as follows: "東京都港区麻布十番 2 丁目 2-4". (It should be noted that this particular address did not appear anywhere in our training data, although the 9-character prefix did appear six times in various different receipts.) After playing around with various character substitutions and deletions, we realized that the LLM was able to reliably correct most corruptions that had appeared in the confusion matrix that was used to generate the synthetic OCR fine-tuning datasets "ocr1" and "ocr10". To test this hypothesis, we deleted the 8th letter "十" and corrupted three other letters (highlighted in **bold** below) in the original address, based on the substitutions that appeared in the confusion matrix, to give the final corrupted character sequence: "東**泉**都**★**区麻布番 2 丁**目** 2-**¥**". Even with this degree of character corruption, the best fine-tuned LLM was able to return all the correct characters, except the final number, where it generated a "2", instead of a "4" (oddly, this was not a substitution present in the confusion matrix). It is also instructive to note, that the LLM was unable to fix the visually similar substitution of the Japanese character "丁" with the Latin character "T" (and which substitution also does not appear in the confusion matrix), even though the pattern of Japanese addresses including this character appears very frequently in the training data.

It is apparent that the confusion matrix might have a strong bearing on the ability of the LLM to learn to perform spelling-correction. Indeed, the confusion matrix that was used only comprised 552 actual character confusion pairs since it was estimated based on a small amount of real data. In future, we propose to greatly expand the confusion matrix, along with other forms of character corruption, such as random newline character insertions, which occur frequently in our test and validation data due to character mis-alignments caused by distorted receipt images.

## 6 CONCLUSIONS

Language models such as BERT, have previously been shown to perform well on the task of extracting NEs from text. Typically in such applications, BERT is used as a classifier to classify individual tokens in the input text, or to classify a span of tokens, as belonging to one of a set of possible NE categories.

In this paper, we fine-tuned eight different LLMs, as well as two different BERT LMs for comparison, on the task of extracting NEs from text that was obtained by applying Optical Character Recognition (OCR) to images of Japanese shop receipts. We showed that the best performing fine-tuned LLM outperformed our best BERT LM on all NE categories except one.

Moreover, in a few cases the best fine-tuned LLM demonstrated the ability to correct the surface form where the input contained OCR errors, as had been hypothesised initially. However, this ability was found to be mostly limited to correcting examples of character errors that the model had been fine-tuned on. More experiments are needed to determine to what extent the spelling-correction ability is inherited from "knowledge" contained in the original, pretrained LLM, versus what it was exposed to during fine-tuning.

## 7 FUTURE WORK

*Spell correction of input text followed by NE extraction.* LLMs used generatively have been shown to perform well "out-of-the-box" at doing spelling correction. Rather than relying on the LLM to perform both NE extraction and spelling correction simultaneously, it would be interesting to compare the performance of NE extraction after using an LLM to spell-correct the input text.

*Prompt Format.* In this work, we intentionally chose a prompt that would work with all the LLMs under investigation. However, this prompt is not necessarily the optimal prompt for each individual LLM. Moreover, preliminary investigations have shown that the instruction-tuned variants of the base LLMs considered in this paper give a consistently higher performance over the respective base LLM variant. For the instruction-tuned LLMs, the optimal choice of prompt is likely to be of even greater importance.

*Data Expansion.* The data that we used typically has multiple annotated ground-truth NEs of the same NE category for a given receipt. However, in this work, for each NE category we only used the first NE that occurred in the training data as the target. (Note however, that when we score an NE produced by an LM, we do actually try to match against all the valid ground-truth NE candidates.) It is possible that better results might be obtained by including all alternative correct NEs as additional training examples. (Of course, it is equally possible that the additional ambiguity from doing this, might make the performance worse.)

In addition, neither the "ocr1" nor the "ocr10" training datasets included the uncorrupted "truth" data. An investigation of whether adding this data is helpful should also be conducted.

*LoRA Parameters.* We have only investigated one set of LoRA parameters in this work, but it is quite likely that different parameter settings would produce different results. Indeed, full parameter fine-tuning of the BERT LMs shows that full parameter fine-tuning of LLMs might also be beneficial, and possibly even better than using LoRA.

*Token Generation Parameters.* There are several parameters which directly affect the generation of tokens, such as *temperature*, *top_k*, and *max_new_tokens*. In this work, none of these have been optimized, yet they can clearly affect the quality of the tokens that are generated. Further optimization of these parameters is likely to be beneficial.

*More Iterations.* One of the LLMs investigated showed optimal performance on the validation data at 10,000 iterations. This suggests that, at least in some cases, it might be beneficial to run more than 10,000 iterations of fine-tuning. Moreover, we should determine whether it would be worth running more iterations on the "ocr10" data in particular, since 10,000 iterations effectively only constitutes just over one epoch over the whole of the "ocr10" data.

*More NE categories.* Preliminary investigations have shown that the multilingual BERT LM performs even better when trained on all 344 NE categories for which we have manually annotated data available. This might be due to beneficial discriminative effects caused by more categories disambiguating semantically similar NEs. Moreover, training on the full set of 344 NE categories would be essential for a production system. Therefore, a proper comparison of all the LLMs, using all 344 NE categories, should be made to determine whether they can reliably generate NEs for a much larger number of fine-grained NE categories. Poor performance of an LLM fine-tuned on a large number of NE categories would obviously be an important factor in deciding whether it makes sense to employ LLMs for NE extraction at all.

## ACKNOWLEDGMENTS

The authors would like to thank Atsushi Kojima, Dr. Hans Dolfing and Professor Ryo Nagata for their feedback on this paper, our 2023 Summer intern, Ryoma Kumon, for the initial code development and data preparation scripts, and Noboru Kubota for his assistance in analyzing the results.